\documentclass[table]{opt2022} 

\usepackage[utf8]{inputenc} 
\usepackage[T1]{fontenc}    
\usepackage{hyperref}       
\usepackage{url}            
\usepackage{booktabs}       
\usepackage{amsfonts}       
\usepackage{amsmath}
\usepackage{wasysym}
\usepackage{bm}
\usepackage{caption}
\newcounter{equationwotag}
\newcommand{\owntag}[2][]{\ifthenelse{\equal{#1}{}}{
    \stepcounter{equation}\tag{\theequation, #2}
}{
    \setcounter{equationwotag}{\value{equation}}\label{#1}\stepcounter{equation}\tag{\theequation, #2}
}}

\title[Counterfactual Explanations Using Optimization With Constraint Learning]{Counterfactual Explanations Using \\ Optimization With Constraint Learning}

\optauthor{%
\Name{Donato Maragno} \Email{d.maragno@uva.nl}\\
\Name{Tabea E. Röber} \Email{t.e.rober@uva.nl}\\
\Name{{\c{S}}. {\.{I}}lker Birbil} \Email{s.i.birbil@uva.nl}\\
\addr Amsterdam Business School, University of Amsterdam, The Netherlands}

\begin{document}






\newcommand{\red}[1]{\textcolor{red}{#1}}
\newcommand{\blue}[1]{\textcolor{blue}{#1}}
\newcommand{\magenta}[1]{\textcolor{magenta}{#1}}
\newcommand{\sibm}[1]{\mbox{\boldmath{$#1$}}}
\newcommand{\rb}[1]{\raisebox{-1.5ex}[0cm][0cm]{#1}}
\newcommand{\HRule}{\noindent\rule{\linewidth}{0.5mm}}
\newcommand{\dsum}{\displaystyle\sum}
\newcommand{\veps}{\varepsilon}

\newcommand{\CA}{\mathcal{A}}
\newcommand{\CB}{\mathcal{B}}
\newcommand{\CC}{\mathcal{C}}
\newcommand{\CD}{\mathcal{D}}
\newcommand{\CG}{\mathcal{G}}
\newcommand{\CI}{\mathcal{I}}
\newcommand{\CJ}{\mathcal{J}}
\newcommand{\CK}{\mathcal{K}}
\newcommand{\CL}{\mathcal{L}}
\newcommand{\CN}{\mathcal{N}}
\newcommand{\CP}{\mathcal{P}}
\newcommand{\CS}{\mathcal{S}}
\newcommand{\CT}{\mathcal{T}}
\newcommand{\CX}{\mathcal{X}}
\newcommand{\ZZ}{\mathbb{Z}}
\newcommand{\RR}{\mathbb{R}}
\newcommand{\NN}{\mathbb{N}}

\newcommand{\va}{\sibm{a}}
\newcommand{\vb}{\sibm{b}}
\newcommand{\vc}{\sibm{c}}
\newcommand{\vd}{\sibm{d}}
\newcommand{\ve}{\sibm{e}}
\newcommand{\vf}{\sibm{f}}
\newcommand{\vg}{\sibm{g}}
\newcommand{\vp}{\sibm{p}}
\newcommand{\vr}{\sibm{r}}
\newcommand{\vs}{\sibm{s}}
\newcommand{\vt}{\sibm{t}}
\newcommand{\vu}{\sibm{u}}
\newcommand{\vv}{\sibm{v}}
\newcommand{\vw}{\sibm{w}}
\newcommand{\vx}{\sibm{x}}
\newcommand{\vy}{\sibm{y}}
\newcommand{\vz}{\sibm{z}}
\newcommand{\zv}{\sibm{0}}
\newcommand{\ov}{\sibm{1}}

\newcommand{\veta}{\sibm{\eta}}
\newcommand{\vxi}{\sibm{\xi}}
\newcommand{\valpha}{\sibm{\alpha}}
\newcommand{\vbeta}{\sibm{\beta}}
\newcommand{\vgamma}{\sibm{\gamma}}
\newcommand{\vtheta}{\sibm{\theta}}
\newcommand{\vlambda}{\sibm{\lambda}}
\newcommand{\vnu}{\sibm{\nu}}
\newcommand{\vmu}{\sibm{\mu}}
\newcommand{\vdelta}{\sibm{\delta}}

\newcommand{\mA}{\sibm{A}}
\newcommand{\mB}{\sibm{B}}
\newcommand{\mC}{\sibm{C}}
\newcommand{\mD}{\sibm{D}}
\newcommand{\mE}{\sibm{E}}
\newcommand{\mF}{\sibm{F}}
\newcommand{\mG}{\sibm{G}}
\newcommand{\mH}{\sibm{H}}
\newcommand{\mI}{\sibm{I}}
\newcommand{\mL}{\sibm{L}}
\newcommand{\mM}{\sibm{M}}
\newcommand{\mP}{\sibm{P}}
\newcommand{\mQ}{\sibm{Q}}
\newcommand{\mR}{\sibm{R}}
\newcommand{\mS}{\sibm{S}}
\newcommand{\mU}{\sibm{U}}
\newcommand{\mV}{\sibm{V}}
\newcommand{\mX}{\sibm{X}}

\newcommand{\tr}{^{\intercal}}
\newcommand{\ntr}{^{-\intercal}}
\newcommand{\inv}{^{-1}}

\newcommand{\gr}{\mbox{graph}}
\newcommand{\ra}{\rightarrow}
\newcommand{\la}{\leftarrow}
\newcommand{\Ra}{\Rightarrow}
\newcommand{\rra}{\rightrightarrows}
\newcommand{\ptr}{\marginpar{$\Leftarrow$}}

\newcommand{\pfxi}{\frac{\partial f(\vx)}{\partial x_i}}
\newcommand{\pfx}{\partial f(\vx)}
\newcommand{\pf}{\partial f}
\newcommand{\pxi}{\partial x_i}
\newcommand{\px}{\partial x}

\newcommand{\nfx}{\nabla f(\vx)}
\newcommand{\eps}{\epsilon}
\newcommand{\eg}{\textit{e.g.}}
\newcommand{\ie}{\textit{i.e.}}

\newcommand{\vsp}{\vspace{4mm}}
\newcommand{\vspp}{\vspace{8mm}}
\newcommand{\vsppp}{\vspace{12mm}}

\newcommand{\hsp}{\hspace{4mm}}
\newcommand{\hspp}{\hspace{8mm}}
\newcommand{\hsppp}{\hspace{12mm}}

\newcommand{\pr}[1]{\mathbb{P}\left\{#1\right\}}
\newcommand{\ex}[1]{\mathbb{E}\left[#1\right]}
\newcommand{\variance}[1]{\mbox{Var}\left(#1\right)}
\newcommand{\covar}[1]{\mbox{Cov}\left(#1\right)}
\newcommand{\C}[2]{\left(\begin{array}{c} #1 \\ #2 \end{array}\right)}

\newcommand{\maximize}{\mbox{maximize\hspace{4mm} }}
\newcommand{\minimize}{\mbox{minimize\hspace{4mm} }}
\newcommand{\subto}{\mbox{subject to\hspace{4mm}}}

\newenvironment{SIBitemize}{
  \renewcommand{\labelitemi}{$\diamond$}
  \begin{itemize}
    \setlength{\parskip}{0mm}}
  {\end{itemize}}





\newcommand{\propnum}[2]{\vspace{3mm}
  \noindent {\sc Proposition #1}{\it #2} \vspace{3mm}}
\newcommand{\lemnum}[2]{\vspace{3mm}
  \noindent {\sc Lemma #1}{\it #2} \vspace{3mm}}
\newcommand{\thmnum}[2]{\vspace{3mm}
  \noindent {\sc Theorem #1}{\it #2} \vspace{3mm}}
\newcommand{\intodo}[2][]{\todo[inline, noshadow, caption={emptytext}, #1]{
    \begin{spacing}{1.0}\normalsize{#2}\end{spacing}}}
    
\maketitle

\begin{abstract}%
To increase the adoption of counterfactual explanations in practice, several criteria that these should adhere to have been put forward in the literature. We propose counterfactual explanations using optimization with constraint learning (\texttt{CE-OCL}), a generic and flexible approach that addresses all these criteria and allows room for further extensions. Specifically, we discuss how we can leverage an optimization with constraint learning framework for the generation of counterfactual explanations, and how components of this framework readily map to the criteria. We also propose two novel modeling approaches to address data manifold closeness and diversity, which are two key criteria for practical counterfactual explanations. We test \texttt{CE-OCL} on several datasets and present our results in a case study. Compared against the current state-of-the-art methods, \texttt{CE-OCL} allows for more flexibility and has an overall superior performance in terms of several evaluation metrics proposed in related work.  
\end{abstract}


\section{Introduction}\label{sec:introduction}

Interpretability in machine learning (ML) is an ongoing research field that has received increasing attention in recent years. Off the many approaches and tools for interpretability, counterfactual explanations (CEs) are expected to be especially promising due to their resemblance to how we provide explanations in everyday life \citep{miller2018}. 
It has been established that we do not seek to explain the cause of an event \textit{per se}, but \textit{relative} to some other event that did not occur. Typically, we have a factual instance vector $\hat{x}$ for which the (prediction) outcome $\hat{y}$ relative to some other, desired, outcome $\tilde{y}$ 
should be explained. The key idea for generating a CE is to find a data point $\bm{\tilde{x}}$ close to the factual instance $\hat{\bm{x}}$, such that the prediction outcome for $\bm{\tilde{x}}$ is $\tilde{y}$.
The difference in the features constitutes the explanation. As CEs do not try to explain all possible causes of an event but focus on necessary changes to the environment to reach a certain state, they tend to be simpler, and with that, also easier to understand than those methods which communicate explanations based on the entire feature space \citep[][]{miller2018}.

\citet{Wachter.2018} are the first to propose an optimization-based approach for generating CEs.  Having a trained classifier $h(\cdot)$, the aim is to find at least one CE, say $\bm{\tilde{x}}$, which has the closest distance to the original factual instance $\hat{\bm{x}}$ such that $h(\bm{\tilde{x}})$ is equal to a different target $\tilde{y}$. Such a CE can be obtained by solving the following mathematical optimization model:
\begin{align}
\label{eqn:wachtermodel}
    \min_{\bm{x}} \max_{\lambda} \lambda (h(\bm{x}) - \tilde{y})^2 + d(\hat{\bm{x}}, \bm{x}),
\end{align}
where $d(\cdot,\cdot)$ is a distance function and $\lambda$ acts as a nonnegative balancing weight to ensure $h(\bm{x}) = \tilde{y}$. Much work has been devoted to refine this problem such that the generated CEs are useful and attainable in practice. From the literature \citep[e.g.,][]{Verma.2020, Wachter.2018, Russell.2019, Navas-Palencia.2021, Mothilal.2020}, we can identify the following eight criteria that a generated CE should fulfill in both theory and practice: \textbf{Proximity:} The CE should be as close as possible to the factual instance $\hat{\bm{x}}$ with respect to the feature values.
\textbf{Validity:} The prediction for the CE $\bm{\tilde{x}}$ should be equal to $\tilde{y} \neq \hat{y}$.
\textbf{Coherence.} When one-hot encoding is used for categorical data, we should be able to map it back to the input feature space to obtain coherent explanations.
\textbf{Sparsity:} The CE should differ from the factual instance in as few features as possible.
\textbf{Actionability:} We can distinguish between immutable, mutable but not actionable, and actionable features. \textbf{Data manifold closeness:} To ensure the generation of realistic and actionable explanations, the generated CEs should be close to the observed (training) data. \textbf{Causality:} Any (known) causal relationships in the data should be respected in the proposed CEs to further ensure realistic explanations. \textbf{Diversity:} Any algorithm for the generation of CEs should return a set of CEs which differ in at least one feature.

These criteria have been partially addressed in recent work, see Table~\ref{tab:overview}. For example, \citet{Russell.2019} and \citet{Ustun.2019} address coherence and actionability, the latter introducing the notion of immutable, conditionally immutable and mutable features. Further, \citet{Russell.2019} focuses on diversity and suggests adding constraints greedily by restricting the state of variables altered in previously generated CEs, while \citet{Mothilal.2020} base their approach to diverse CEs on determinantal point processes \citep{Kulesza_2012}. \citet{Kanamori.2020} attempt to optimize the idea of proximity and data manifold closeness using \textit{Mahalanobis’ distance} and the \textit{local outlier factor} to generate CEs close to the empirical distribution of the training data. \citet{Poyiadzi2020} base their work on graph theory, and apply a shortest path algorithm to minimize the $f$-distance quantifying the trade-off between the path length and the density along this path, by that ensuring a solution that lies in a high density region. To address causality, \citet{Kanamori.2020jn} discuss the use of a structural causal model (SCM), while others advocate a post-hoc filtering approach \citep{Mothilal.2020}. We refer to \citet{Verma.2020} and \citet{Guidotti2022} for an extensive overview of recent works on counterfactual explanations.

To the best of our knowledge ours is the first work that addresses all of these criteria in a combined setting 
. We propose \texttt{CE-OCL}, a generic and flexible approach for generating CEs based on optimization with constraint learning (OCL). OCL is a new and fast-growing research field whose aim is to learn parts of an optimization model (\textit{e.g.}, constraints or objective function) using ML models whenever explicit formulae are not available (see \citet{fajemisin2021optimization} for a recent survey on OCL). We show how all the criteria proposed in the literature can be addressed by an OCL framework. Based on the concept of trust regions, we also propose a new modeling approach to ensure data manifold closeness and coherence. Finally, we propose using incumbent solutions to obtain diverse CEs in a single execution. With our extensive demonstration on standard datasets from the CE literature, we also set new benchmarks for future research.

\renewcommand{\baselinestretch}{1.1} 

\begin{table}[ht]
\caption{State-of-the-art methods to generate CEs}
\label{tab:overview}
\resizebox{\textwidth}{!}{%
\begin{tabular}{@{}lccccccc@{}}
\toprule

& \textbf{Proximity} & \textbf{Sparsity} & \textbf{Coherence} & \textbf{Actionability} & \textbf{Data Manifold Closeness} & \textbf{Causality} & \textbf{Diversity} \\ \midrule
\rowcolor{gray!10} \citet{Laugel2017} & \CIRCLE & \CIRCLE & \CIRCLE & -- & -- & -- & -- \\ 
\citet{Russell.2019}  & \CIRCLE & \LEFTcircle & \CIRCLE & -- & -- & -- & \CIRCLE \\
\rowcolor{gray!10}\citet{Ustun.2019} &  \CIRCLE & \CIRCLE & \CIRCLE & \CIRCLE & -- & -- & -- \\
\citet{Kanamori.2020} & \CIRCLE & -- & \CIRCLE & -- & \CIRCLE & -- & -- \\
\rowcolor{gray!10}\citet{Mahajan.2019} & \CIRCLE & -- & \CIRCLE & \LEFTcircle & \CIRCLE & \CIRCLE & -- \\
\citet{Karimi.2020} & \CIRCLE & -- & \CIRCLE & -- & -- & \CIRCLE & -- \\
\rowcolor{gray!10}\citet{Kanamori.2020jn} & \CIRCLE & \CIRCLE & \CIRCLE & \CIRCLE & -- & \CIRCLE & \CIRCLE \\
\citet{Mothilal.2020} & \CIRCLE & \LEFTcircle & \CIRCLE & \CIRCLE & -- & \LEFTcircle &\CIRCLE \\
\rowcolor{gray!10}\citet{Karimi.2019fy} & \CIRCLE & \CIRCLE & \CIRCLE & \CIRCLE & -- & -- &\CIRCLE \\
\citet{Poyiadzi2020} & \CIRCLE & -- & \CIRCLE & \CIRCLE & \CIRCLE & -- & -- \\
\rowcolor{gray!10}\texttt{CE-OCL} & \CIRCLE & \CIRCLE & \CIRCLE & \CIRCLE & \CIRCLE & \CIRCLE & \CIRCLE \\ \bottomrule
\multicolumn{8}{l}{\footnotesize \CIRCLE: addressed; \LEFTcircle: partially addressed; --: absent}

\end{tabular}}
\end{table}

\section{Generation of counterfactual explanations}\label{sec:generate}

In an OCL framework, ML models are used to design constraint and objective functions of an optimization model when explicit expressions are unknown. First, the predictive model is trained on historical data and then it is embedded into the optimization model using decision variables as inputs \cite{Biggs2017RF, verwer2017auction, villarrubia2018artificial}. Although the interplay between optimization and ML has a different aim in OCL than CE generation, we notice that the two frameworks have a similar structure. Recent advances in OCL successfully reduce the computational burden of embedding fitted ML models into an optimization model \cite{Grimstad_2019, Schweidtmann2019, Misic2020} and can be easily transferred to the problem of generating CEs.
In this regard, we show how the problem of generating CEs, given a fitted model $h(\cdot)$, a factual instance $\bm{\hat{x}}$, and the desired outcome $\tilde{y}$, can be seen as a special case of \textit{optimization with constraint learning}. 
In an OCL setting, a dataset $\mathcal{D} = \{(\bm{\bar{x}_i}, \bar{y}_i)\}_{i=1}^N$ with observed feature vector $\bm{\bar{x}_i}$ and  outcome of interest $\bar{y}_i$ for sample $i$, is used to train predictive models that are to be constrained or optimized in a larger optimization problem. An OCL model is typically presented as
\begin{subequations}
\begin{align}
    \underset{{\bm{x}\in \mathbb{R}^n,y\in \mathbb{R}}}{\minimize} \ & f(\bm{x}, y) & \label{eqn:conceptualmodelCL1}\\
    \subto \ & \bm{g}(\bm{x}, y) \leq \bm{0}, & \label{eqn:conceptualmodelCL2}\\
    & y = h(\bm{x}), & \label{eqn:conceptualmodelCL3}\\
    & \bm{x} \in \mathcal{X},& \label{eqn:conceptualmodelCL4}
\end{align}
\end{subequations}
where $\bm{x} \in \mathbb{R}^n$ is the decision vector with components $x_i \in \RR$, $f(\cdot,\cdot):\mathbb{R}^{n+1} \mapsto \mathbb{R}$ and $\bm{g}(\cdot,\cdot):\mathbb{R}^{n+1} \mapsto \mathbb{R}^m$ are known functions possibly also depending on the predicted outcome $y$, and $h(\cdot):\mathbb{R}^{n} \mapsto \mathbb{R}$ represents the predictive model\footnote{To simplify our exposition, we include only one predictive model. However, a general OCL framework admits multiple learned constraints in the model.} trained on $\mathcal{D}$. The set $\mathcal{X}$ defines the trust region, \textit{i.e.}, the set of solutions for which we trust the embedded predictive models (see below for details).
Formulation (\ref{eqn:conceptualmodelCL1}-\ref{eqn:conceptualmodelCL4}) is quite general and encompasses a large body of work that includes CE generation. Now, we characterize the parallelism between some of the eight criteria listed in Section~\ref{sec:introduction} and the structure of the resulting OCL model. We elaborate and discuss the remaining criteria in Appendix~\ref{app:generate}.

\paragraph{Validity.} While the trained model $h(\cdot)$ is used in constraint learning to define, completely or partially, the objective function and/or the constraints, in CE generation it is used to enforce the validity constraint. Constraint (\ref{eqn:conceptualmodelCL3}) is likely to be an encoding of the predictive model. In other words, embedding a trained ML model requires adding multiple constraints and auxiliary variables. When $h(\cdot)$ is a classification model, the CE validity is obtained by constraining the model prediction to be equal to the desired class $\tilde{y}$; that is, we set $y=\tilde{y}$. If $h(\cdot)$ is a regression model, the OCL framework still applies, and an inequality constraint can be used to enforce validity; \textit{e.g.}, $y\leq\tilde{y}-\delta$ or $y\geq\tilde{y}+\delta$ for some fixed $\delta \in \mathbb{R}_+$.

\paragraph{Data manifold closeness.} One of the requirements to obtain plausible CEs is that they are close to the data manifold. For this purpose, we can make use of the \textit{trust region} constraints. \citet{maragno2021mixedinteger} define the trust region as the convex hull (CH) of $\mathcal{D}$ in the features space, and they use it in OCL to prevent the trained model from extrapolating, therefore, mitigating the deterioration in predictive performance for points that are farther away from the data points in $\mathcal{D}$. In CE generation, the trust region, or rather \textit{data manifold region}, serves the purpose of ensuring solutions in a high-density region. To this end, we can also denote a CE ($\bm{\tilde{x}}$) as the convex combination of samples in $\mathcal{D}$, in particular samples belonging to the desired class ($\tilde{y}$). 

In case the CH is too restrictive, we can use a relaxed formulation to enlarge the data manifold by including those solutions that are in the $\epsilon$-ball surrounding some feasible solutions in the CH:
\begin{align}
\text{$\epsilon$-CH} = \bigg\{ \bm{x} \bigg| \sum_{i \in \mathcal{I}} \lambda_i \bm{\bar{x}}_i = \bm{x} + \bm{s}, \ \sum_{i \in \mathcal{I}} \lambda_i = 1, \ \bm{\lambda} \geq 0, \ ||\bm{s}||_{p} \leq \epsilon
    \bigg\},
\label{eqn:epstrustregionconstr} 
\end{align}
where $\lambda_i \in [0,1]$ and $\bm{s} \in \mathbb{R}^n$ are auxiliary variables, $\epsilon \geq 0$ is a hyperparameter, and $\mathcal{I}$ denotes the indices corresponding to the subset of samples in $\mathcal{D}$ belonging to the desired class $\tilde{y}$. When $\epsilon=0$, we obtain the trust region as discussed in \citet{maragno2021mixedinteger}. However, $\epsilon > 0$ leads to a less restrictive set of conditions. Further details are available in Appendix~\ref{app:generate} as well as a graphical representation of the data manifold closeness in Figure~\ref{fig:test}.

\paragraph{Causality.}
CEs might be inefficient or unrealistic when causal relations are not considered in the generation process. Both these situations are exemplified in \citet{Karimi.2020}, where the authors show the importance of causal relations to obtain CEs that better answer the question 
“what \textit{should be done} in the future considering the laws governing the world.” When a causal model is available, we can formulate the causal relations among variables as extra constraints of the optimization model. When there is not an explicit formulation of the causal relations, we are in a typical constraint learning scenario where an ML model can be trained and embedded into the optimization. We provide the formulation of causality constraints in Appendix~\ref{app:generate}.

\paragraph{Diversity.} Most of the methods for generating multiple and diverse CEs in the literature require multiple runs and extra constraints to generate diverse CEs for the same input. Following an iterative approach, we can generate diverse CEs using constraints on the actionability of features \citep{Russell.2019}, or constraints on the distance between the subsequent CE and all the previously generated ones \citep{Karimi.2019fy}. Again in an iterative way, we can also use the data manifold constraints to generate diverse CEs (i) by finding one CE for each clustered CH, (ii) by enlarging the CH with increasing $\epsilon$ whenever the data manifold constraints are active.
The use of diversity constraints offers great flexibility at the expense of computation time. As an alternative, we propose to solve one single optimization model and use the pool of \textit{incumbent solutions} as the set of CEs. In mixed-integer optimization, solvers like Gurobi or CPLEX allow retrieving the sub-optimal solutions found during the tree search procedure \citep{gurobi, cplex2009v12}. In this way, collecting a set of CEs comes at no cost in terms of computation time.

\section{Experiments and results}\label{sec:experiments}
\setlength{\textfloatsep}{10pt plus 1.0pt minus 2.0pt}
In this section, we demonstrate the effectiveness of OCL through empirical experiments on multiple datasets and comparing the results with other state-of-the-art methods. The experiments are executed using \texttt{OptiCL} \footnote{\hyperlink{https://github.com/hwiberg/OptiCL}{https://github.com/hwiberg/OptiCL}, under the MIT license.} \citep{maragno2021mixedinteger}, an open-source Python package for optimization with constraint learning. \texttt{OptiCL} has been originally designed to help practitioners in modeling an optimization problem whose constraints are partially unknown, but where ML models can be deployed to learn them \citep{maragno2021mixedinteger}. However, as detailed in Section~\ref{sec:generate}, the problem of generating CEs directly relates to an OCL problem. \texttt{OptiCL} currently supports several MIO-representable predictive models, including logistic regression (lr), support vector machines (svm), (optimal) decision trees (cart), random forests (rf), gradient boosting machines (gbm), and neural networks with ReLU activation functions (mlp). Moreover, \texttt{OptiCL} allows for trust region constraints as defined in (\ref{eqn:epstrustregionconstr}). Whenever a causal model is available but the relations are not explicit, \texttt{OptiCL} allows representing the relation using one of the MIO-representable ML models. The open-source implementation for reproducing all our results is available at \texttt{https://github.com/tabearoeber/CE-OCL}.

We performed an extensive comparison of our method against four state-of-the-art methods: \texttt{Growing Spheres (GS)} \cite{Laugel2017}, \texttt{FACE}\cite{Poyiadzi2020}, \texttt{Actionable Recourse (AR)} \cite{Ustun.2019}, and \texttt{DiCE} \cite{Mothilal.2020}. Here, we present the results on the COMPAS dataset\footnote{Appendix~\ref{app:comparison} includes the results for three further datasets: Adult, Give Me Some Credit, and HELOC.}. 
We generated CEs for 30 factual instances, then averaged the scores on the evaluation metrics proposed by \citet{Mothilal.2020}. Our results are presented in Figure \ref{fig:compas_results} (see Appendix \ref{app:comparison} for details on the evaluation metrics). For the sake of clarity, we have rescaled values such that they range from 0 (worst) to 1 (best).
Since the majority of the other methods do not generate a set of CEs, we chose to generate only one CE for each factual instance and hence do not report any diversity scores (Figure \ref{fig:compas_results} left). Furthermore, we compare our approach in terms of diversity by generating three CEs for each of the 30 factual instances and compare the results with \texttt{DiCE} \citep{Mothilal.2020} (Figure \ref{fig:compas_results} right). Extensive results for all datasets and with different predictive models are included in Appendix \ref{app:comparison}.
\begin{figure}
    \centering
    \includegraphics[width=0.49\textwidth]{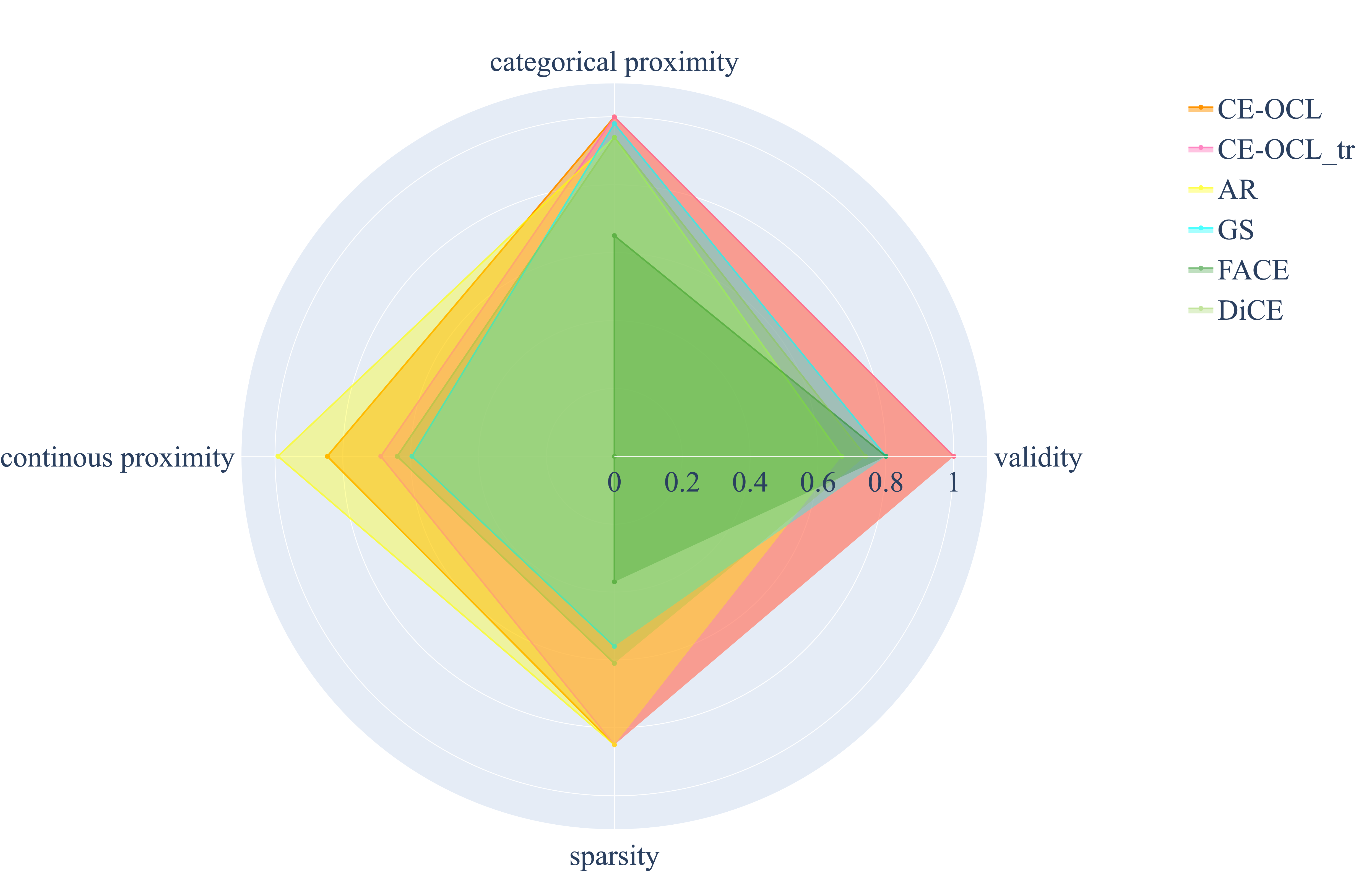}
    \includegraphics[width=0.49\textwidth]{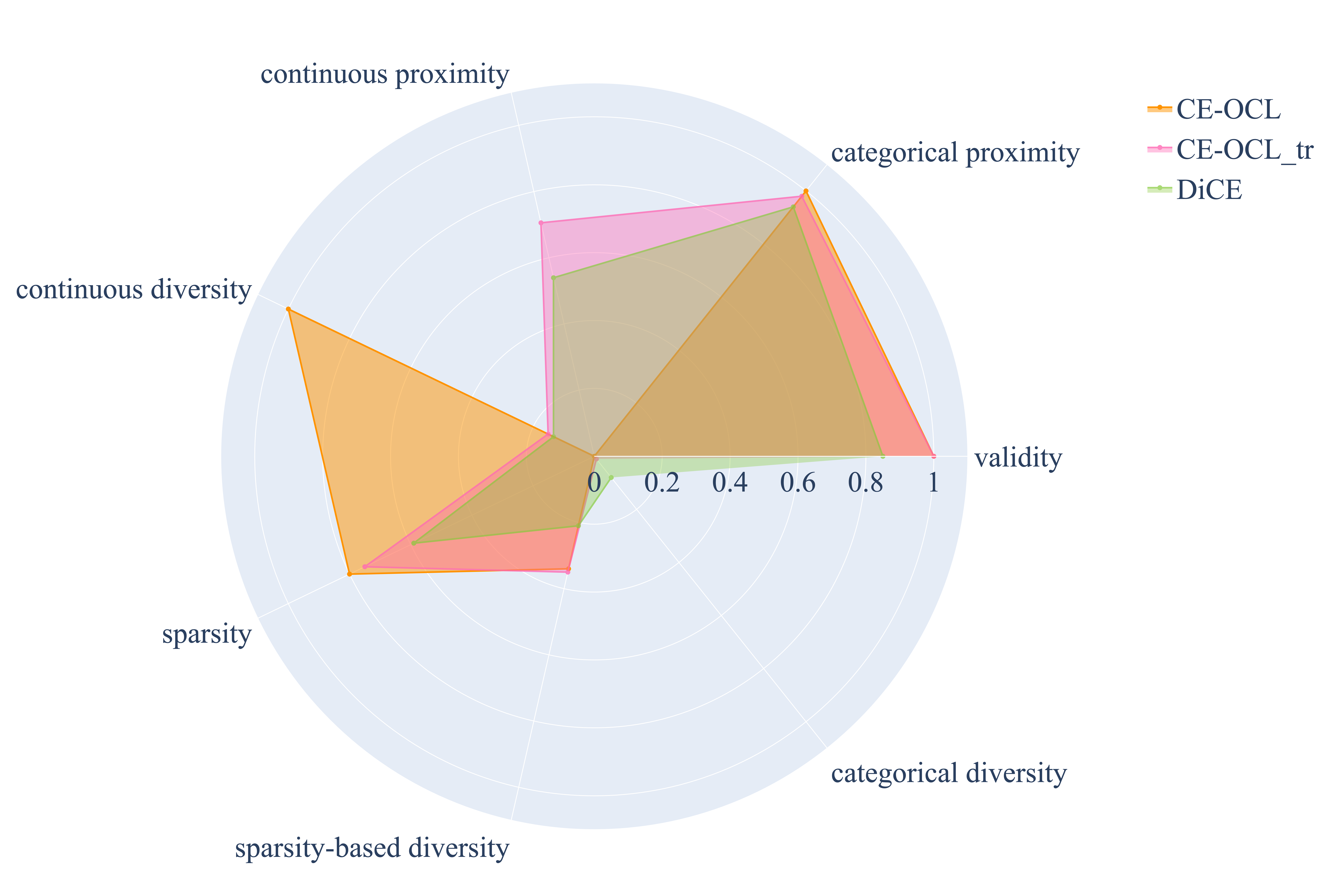}
    \caption{\footnotesize Performance of \texttt{CE-OCL} and \texttt{CE-OCL\_tr} (with trust region) compared to performance of current state-of-the-art methods for generating counterfactual explanations on the COMPAS dataset. Left: we generated one counterfactual for each of 30 factual instances; rf as predictive model. Right: we generated three counterfactuals for the same instances; lr as predictive model. Not all methods support generating several counterfactuals.}
    \label{fig:compas_results}
\end{figure}
We further demonstrate the generation of CEs in a step-wise manner on the Statlog (German Credit Data) dataset \citep[][]{Dua:2019}, which is one of the standard datasets in the CE literature\footnote{We also provide another demonstration on the Statlog (Heart) dataset \citep[][]{Dua:2019} in Appendix~\ref{app:case study}.}. The German Credit dataset classifies people described by a set of 20 features as good or bad credit risk, see Table \ref{tab:description} in Appendix~\ref{app:german_tables} for an overview of the features. For this demonstration, we gradually add constraints to the model and present the generated CEs at each step in Table~\ref{tab:credit_demo} shown in Appendix~\ref{app:german_tables}. The table is divided into six parts (A-F), each showing the set of CEs generated, and a dash is used to represent no change to the corresponding features. In Table~\ref{tab:credit_demo} in Appendix~\ref{app:german_tables}, we present the evaluation of these CEs using several evaluation metrics proposed by \citet{Mothilal.2020}: validity, sparsity, categorical and continuous proximity, categorical and continuous diversity, and sparsity-based diversity. The complete mathematical model is detailed in Appendix~\ref{app:german_model}.

We fit several ML models to the data, all of which performed similarly well. For demonstration purposes, we have chosen a linear support vector machine. The factual instance $\bm{\hat{x}}$ used for this case study is reported in Table \ref{tab:credit_demo}.
We start the demonstration considering only validity, proximity, and coherence (Part A), and using the $\ell_2$-norm as a distance function. The optimal solution suggests several changes in the factual instance and is not actionable in practice due to the negative value for F2 (credit amount). To induce sparsity (Part B), we use auxiliary variables to keep track of the number of features changed and penalize them in the objective function. Multiple and diverse CEs are generated using incumbent solutions (Part C). To ensure that the set of generated CEs is valuable in practice, we add actionability constraints  (Part D).
Respecting these constraints, the set of generated CEs seems more realistic however, they may still not be attainable in practice. Specifically, if we consider solution (c) of Part D, the only suggested change concerns F4 (age). However, this CE is unlikely to represent a realistic data point, considering the other feature values remain unchanged. In other words, CEs that do not resemble the training data come with the risk of being unattainable in practice. To this end, we use the idea of a \textit{data manifold region}, as detailed in Section \ref{sec:generate}. As a result, in Part E, we obtain a more realistic set of CEs, although at the expense of sparsity and (categorical) proximity (see the scores reported in Table~\ref{tab:credit_demo}, Appendix~\ref{app:german_tables}). From a qualitative point of view, the three CEs show a more sensible combination of feature values compared to those in Part D. 
Finally, we can leverage the partial SCM provided by \citet{Karimi.2020} for this dataset, which shows that F1 (duration) is causally related to F2 (credit amount). This relationship is learned by a multi-layer perceptron (MLP) using 5-fold cross validation. In Part F, we display the set of CEs that satisfy also the learned causality constraints.

\section{Discussion}\label{sec:conclusion}

With this work, we propose \texttt{CE-OCL}, a generic approach for generating sensible and practical counterfactual explanations. In Section~\ref{sec:experiments}, we report the generally superior performance achieved by \texttt{CE-OCL} compared to other popular methods. 
Nevertheless, we acknowledge the limitations of using incumbent solutions as multiple counterfactuals caused by the lack of control over the solutions' diversity. Whenever we have specific diversity requirements to meet, the iterative approaches proposed by \citet{Russell.2019} and \citet{Karimi.2019fy} may suit best.
Moreover, owing to the MIO structure of \texttt{CE-OCL} and various constraints used to satisfy the established criteria, the feasibility space may shrink to the point of being empty, making the optimization problem infeasible. In the infeasibility case, we recommend following an approach similar to that presented in Section~\ref{sec:experiments}, where constraints are added one at a time. Infeasibility problems due to data manifold constraints can be mitigated by enlarging the data manifold region at the (potential) expense of the sensibility of the CEs. For future research, we plan to investigate the effect of clustering and enlargement of the data manifold region on the CE quality and on diversity. We also intend to extend \texttt{CE-OCL} with additional criteria like robustness in the sense that the generated CEs are not point solutions, but that they are defined by ranges in the feature values.

\section*{Acknowledgments}{This work was supported by the Dutch Scientific Council (NWO) grant OCENW.GROOT.2019.015, Optimization for and with Machine Learning (OPTIMAL).}
\bibliography{main}

\newpage

\appendix

\section{Generate counterfactuals}\label{app:generate}
In Section~\ref{sec:generate}, we describe how criteria such as validity, closeness, causality, and diversity can be fulfilled exploiting OCL components. Likewise, other criteria can be mathematically represented in the following way:
\paragraph{Proximity.} By definition, a CE has to be in the proximity of the factual instance according to some user-defined distance function. To obtain a CE $\bm{\tilde{x}}$ in the proximity of $\bm{\hat{x}}$, we can write the objective function (\ref{eqn:conceptualmodelCL1}) as a distance function $d(\bm{x}, \hat{\bm{x}})$. In the literature, this function is represented by $\ell_1$-norm, $\ell_2$-norm, or as the Mahalanobis’ distance.

\paragraph{Coherence.} When one-hot encoding is used to deal with categorical features, we can use the constraints proposed by \citet{Russell.2019} to obtain coherent CEs. That is, we write for $k$ categorical features the following constraints:
\begin{align}
    \sum_{j' \in \CC_j} x_{j'} = 1, ~~ j = 1,\dots, k,\label{eqn:coherence}
\end{align}
where $\CC_j$ is a set of indices referring to the dummy (binary) variables used to represent the categorical feature $j$.
The use of a data manifold region (with a sufficiently small $\epsilon$) has an interesting impact on CE coherence because constraints (\ref{eqn:coherence}) become redundant. To exemplify how data manifold constraints guarantee coherence, we consider a set of samples represented by the set of indices $\CI$, and a categorical feature \textit{diet} that can assume only three values: \textit{vegan}, \textit{vegetarian}, or \textit{omnivore}. 
We use one-hot encoding to replace the feature \textit{diet} and describe a CE with the dummy (binary) variables $x_{vegan}$, $x_{vegetarian}$, $x_{omnivore}$. From (\ref{eqn:epstrustregionconstr}), we have
\begin{align*}
    x_j = \sum_{i\in \CI} \lambda_i\bar{x}_{i,j},  ~~ j\in \{vegan, vegetarian, omnivore\},
\end{align*}
with $\sum_{i\in\CI} \lambda_i = 1$. One of the dummy variables, say $x_{vegan}$, can assume value 1 only if it is the convex combination of data points $\bar{\vx}_i$ with $\bar{x}_{i,vegan} = 1$ and $\bar{x}_{i,vegeterian} = \bar{x}_{i,omnivore} = 0$. Thus, $\lambda_i > 0$ only when $\bar{x}_{i,vegan}=1$, and consequently, we obtain $x_{vegetarian} = x_{omnivore} = 0$.

\paragraph{Sparsity.} The sparsity can be handled by enforcing the following set of constraints:
\begin{subequations}
\begin{align}
    & \ |x_j - \hat{x}_j| \leq M z_j, ~~ j=1, \dots, n,\label{eqn:sparsity1}\\
    & \ \sum_{i=1}^n z_i \leq K,\label{eqn:sparsity2}
\end{align}
\end{subequations}
where $z_j \in \{0,1\}$, $j=1, \dots, n$ are auxiliary variables that are simply used to count the number of features in $\bm{x}$ that differ from $\hat{\bm{x}}$, and $K$ is an upper bound on the number of allowed changes. Alternatively, constraints (\ref{eqn:sparsity2}) can be relaxed and moved to the objective function with a scaling penalty factor $\alpha > 0$. That is, we obtain the new objective function  $f(\bm{x}, y) + \alpha\sum^n_{i=1} z_i$. Though simpler, this relaxation does not guarantee to lead to an optimal solution with less than or equal to $K$ changes.

\paragraph{Actionability.} As a recommended CE should never change the immutable features, we can restrict the CE to be equal to the factual instance for all the immutable features. Suppose that the set of immutable features is represented by $\mathcal{I}_m$, then we simply add the following constraints:
\begin{align}
     x_i = \hat{x}_i, ~~ i \in \mathcal{I}_m.
\end{align}
Other feasibility constraints might concern actionable variables that cannot take certain values, such as \textit{age}, which can only be increased, or \textit{has\_phd}, which can only change from false to true. These conditions can be added exactly like immutable features.

\paragraph{Data manifold region.}

\begin{figure}
\centering
  \includegraphics[width=0.49\textwidth]{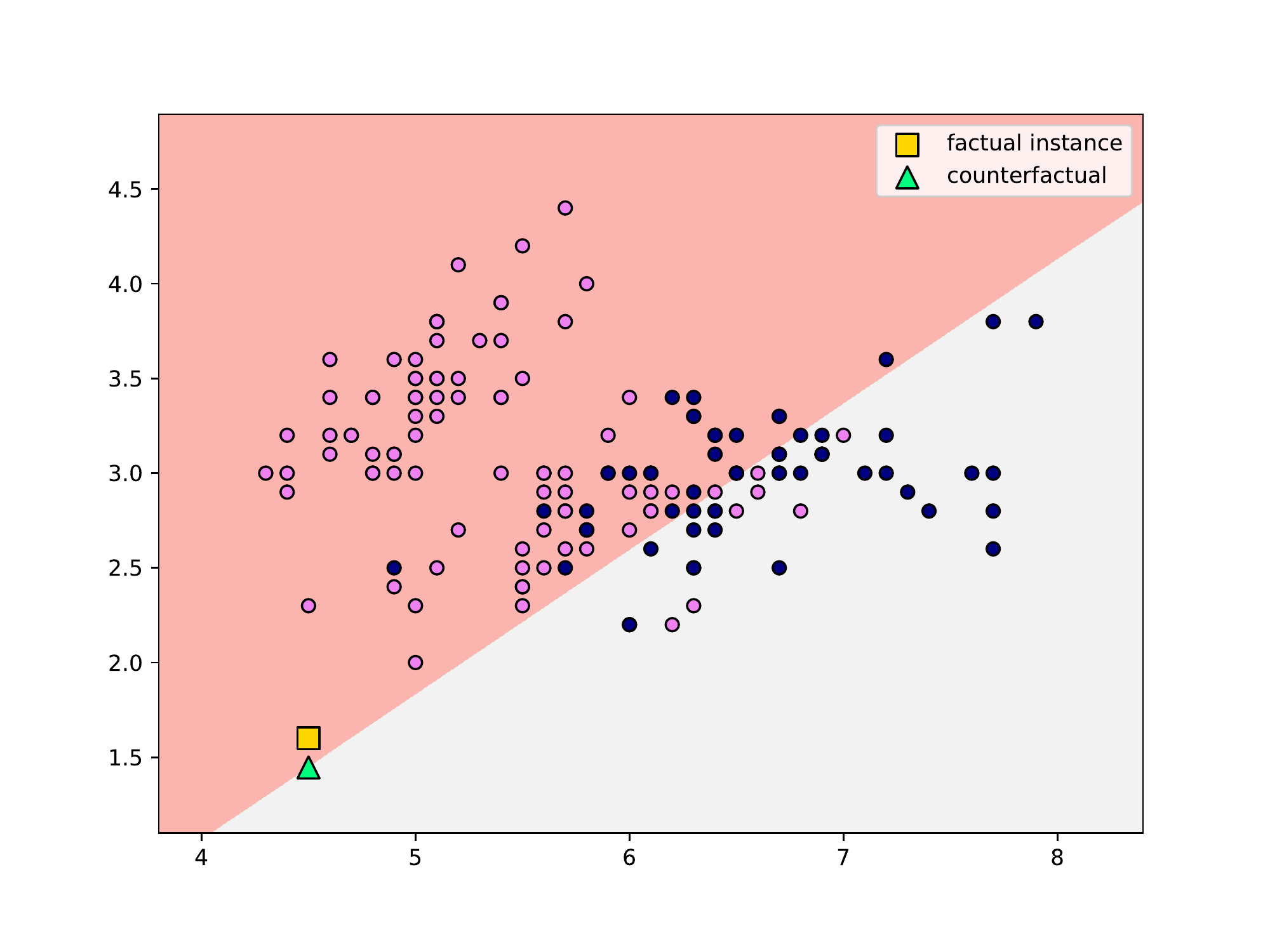}
  \includegraphics[width=0.49\textwidth]{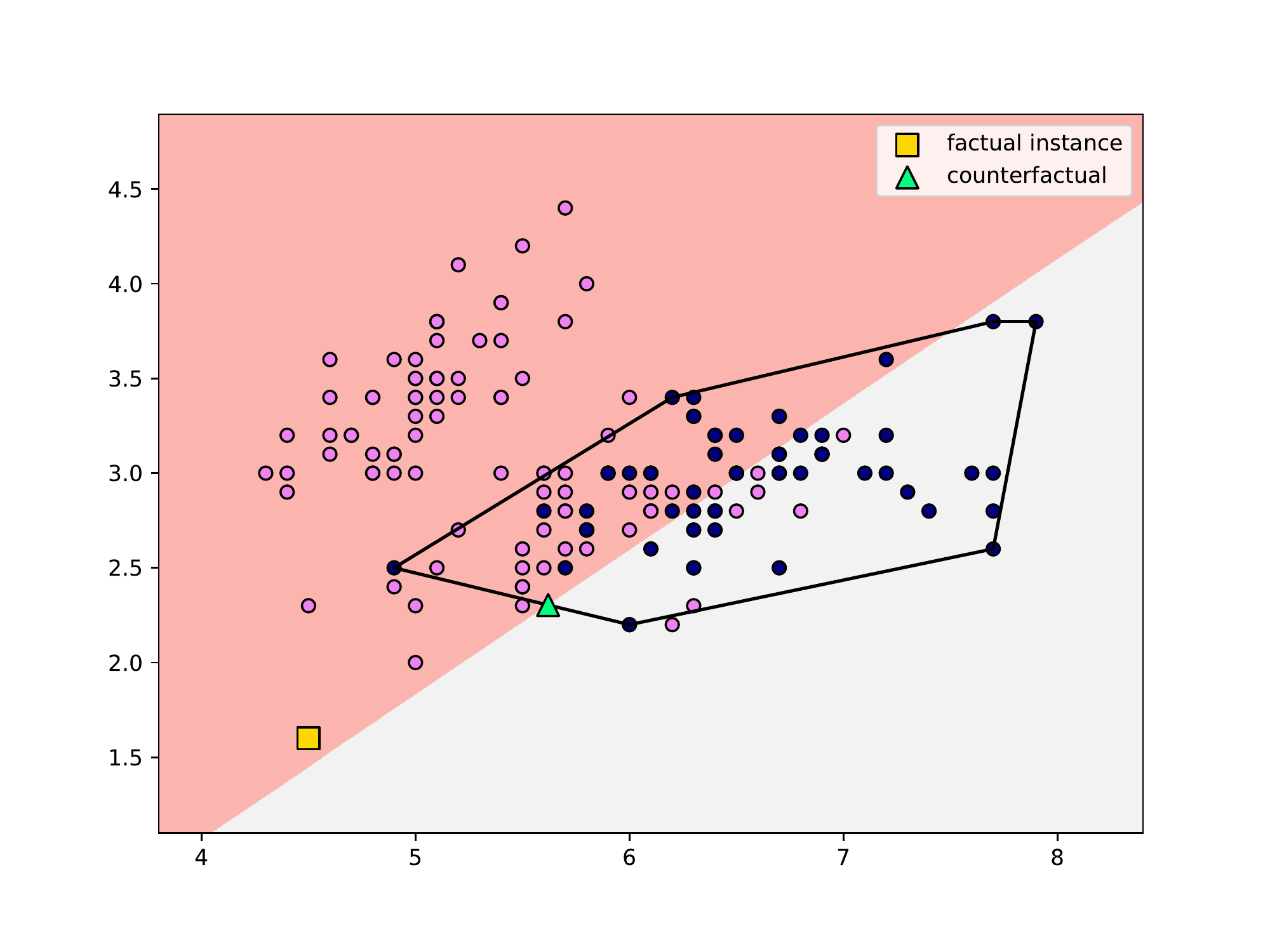}
  \caption{The effect of the data manifold region on the generated CE. The left figure shows the factual instance and its closest counterfactual without closeness constraints. The right figure shows the same factual instance with the CE constrained to be within the data manifold region.}
\label{fig:test}
\end{figure}

Figure~\ref{fig:test} shows how the CH of $\mathcal{D}$ in the features space ensures a solution closer to the data manifold, leading to more plausible CEs. In some cases, the CH may be too restrictive, which is why we introduce formulation~\ref{eqn:epstrustregionconstr} to enlarge the data manifold region by including solutions that are in the $\epsilon$-ball around some feasible solutions in the CH. Being able to enlarge the data manifold region represents a solution to the criticism by \citet{Balestriero2021}: “[...] interpolation\footnote{Interpolation occurs for a sample $\bm{x}$ whenever this sample belongs to the CH of a set of data points.} almost surely never occurs in high-dimensional spaces $(> 100)$ regardless of the underlying intrinsic dimension of the data manifold.”  Aside from the bound on the norm of $\vs$, all constraints in \eqref{eqn:epstrustregionconstr} are linear. Fortunately, the most common norms used to constraint $\bm{s}$ are $\ell_1$-, $\ell_2$-, or $\ell_\infty$-norm. These norms lead to convex conic constraints that can be handled easily with off-the-shelf optimization solvers. The effectiveness of the data manifold region might be hampered by the fact that the CH includes low-density regions. In this case, \citet{maragno2021mixedinteger} advocate a two-step approach: first, clustering is used to identify distinct high-density regions, and then, the data manifold region is represented as the union of the (enlarged) convex hulls of the individual clusters.

\noindent
\textbf{Causality}
The causality constraints are modelled by applying the Abduction-Action-Prediction steps \citep{Pearl.2013}, \citet{Karimi.2020} define the endogenous variables (with indices in the set $\mathcal{E}$) as
\begin{align}
    x_i = \hat{x}_i + c_i(\bm{p}_i) - c_i(\hat{\bm{p}}_i), ~~ i \in \mathcal{E},
\end{align}
where $c_i(\bm{p}_i)$ is a function of the parents of $x_i$, namely the predecessors of the feature $i$ in the SCM. Both $\hat{x}_i$ and $c_i(\hat{\bm{p}}_i)$ are known before the optimization and therefore treated as parameters.
When there is not an explicit formulation of $c_i(\cdot)$, we are in a constraint learning scenario where an ML model can be trained and embedded into the optimization as $c_i = h_i(\bm{p}_i)$ for all $i \in \mathcal{E}$.

\section{Comparison against other methods}\label{app:comparison}

We compared \texttt{CE-OCL} to four open-source tools for generating CEs: \texttt{Growing Spheres} \cite{Laugel2017},  \texttt{FACE}\cite{Poyiadzi2020}, \texttt{Actionable Recourse} \cite{Ustun.2019}, and \texttt{DiCE} \cite{Mothilal.2020}. The experiments are performed using \texttt{CARLA} \cite{pawelczyk2021carla}, a Python library to benchmark counterfactual explanation and recourse models. The predictive model used in the experiments is a random forest and the evaluation is performed by generating a counterfactual for 30 different factual instances on four datasets available in \texttt{CARLA}: Adult, Give Me Some Credit, COMPAS, and HELOC. We average the results for the evaluation metrics proposed by \citet{Mothilal.2020} and present them together with the standard error (s.e.) in Table~\ref{tab:carla_comparison}.
Validity, sparsity, categorical proximity, categorical diversity, and sparsity-based diversity range in the interval [0,1], where 0 and 1 represent the worst and the best scores ($\uparrow^1_0$), respectively. Continuous diversity is a positive number, and the higher it is, the better ($\uparrow_0^+$). Continuous proximity is a negative number, and the closer it is to 0, the better ($\uparrow^0_{-}$).

While \texttt{CE-OCL} can deal with causality and closeness constraints, this does not apply to \texttt{DiCE} which uses a post-hoc filtering approach to remove unrealistic CEs. In addition to causality and closeness constraints, \texttt{Actionable Recourse}, and Growing Sphere cannot generate more than one counterfactual for each instance. \texttt{FACE} does not support diversity and causality constraints but it is able to generate CEs close to the data manifold region. 
Therefore, in Table~\ref{tab:carla_comparison} we report both the results obtained with \texttt{CE-OCL} including validity, proximity, coherence, sparsity, and immutability constraints, and the results obtained including also the closeness constraints, \texttt{CE-OCL\_tr}. The results show that, across all datasets, both \texttt{CE-OCL} and \texttt{CE-OCL\_tr} exhibit better performance in terms of validity, categorical proximity, and sparsity. \texttt{Actionable Recourse} and \texttt{CE-OCL}/\texttt{CE-OCL\_tr} perform equally well in terms of continuous proximity.

\renewcommand{\baselinestretch}{1}
\begin{table}[ht]
\caption{Comparison of CE-OCL with DiCE (genetic), Algorithmic Recourse, Growing spheres, and FACE using Random Forest as predictive model.}
\label{tab:carla_comparison}
\resizebox{\textwidth}{!}{%
\begin{tabular}{@{}llcclc@{}}
\toprule
                              &                           & \textbf{validity} ($\uparrow_0^1$)       & \textbf{cat. proximity} ($\uparrow_0^1$)       & \textbf{cont. proximity} ($\uparrow_-^0$)      & \textbf{sparsity} ($\uparrow_0^1$)      \\
                              &                           & mean (s.e.)                     & mean (s.e.)                                  & mean (s.e.)                                & mean (s.e.)       \\ \midrule
\rowcolor{gray!10}                & CE-OCL                       & \textbf{1.00} (0.00)                        & \textbf{1.00} (0.00)                                    & -4844.35 (575.93)                             & \textbf{0.93} (0.00)                       \\
\rowcolor{gray!10} & CE-OCL\_tr & \textbf{1.00} (0.00) & 0.97 (0.02) & -21785.50 (7506.35) & 0.86 (0.01) \\
\rowcolor{gray!10}                & DiCE                & \textbf{1.00} (0.00)                        & 0.74 (0.03)                                    & -84278.61 (11613.30)                          & 0.50 (0.02)                       \\
\rowcolor{gray!10} ADULT           & Actionable Recourse          & \textbf{1.00} (0.00)                        & 0.78 (0.07)                                    & \textbf{0.00} (0.00)                                   & 0.89 (0.04)                       \\
\rowcolor{gray!10}                & Growing Spheres              & 0.80 (0.07)                        & 0.95 (0.01)                                    & -78901.08 (10395.08)                          & 0.59 (0.01)                       \\
\rowcolor{gray!10}                & FACE                         & 0.80 (0.07)                        & 0.65 (0.03)                                    & -108614.33 (18804.05)                         & 0.47 (0.02)                       \\ 
                & CE-OCL                       & \textbf{1.00} (0.00)                        & \textbf{1.00} (0.00)                                    & -15.23 (5.84)                                 & \textbf{0.85} (0.01)                       \\
                & CE-OCL\_tr & \textbf{1.00} (0.00) & 1.00 (0.00) & -30.83 (16.00) & \textbf{0.85} (0.01) \\
                & DiCE                & 0.74 (0.09)                        & 0.94 (0.03)                                    & -35.58 (9.59)                                 & 0.61 (0.01)                       \\
COMPAS          & Actionable Recourse          & 0.67 (0.11)                        & 0.94 (0.03)                                    & \textbf{-0.87} (0.10)                                  & \textbf{0.85} (0.01)                       \\
                & Growing spheres              & 0.80 (0.07)                        & 0.98 (0.02)                                    & -39.88 (6.58)                                 & 0.56 (0.01)                       \\
                & FACE                         & 0.80 (0.07)                        & 0.65 (0.05)                                    & -98.83 (21.18)                                & 0.37 (0.03)                       \\ 
\rowcolor{gray!10}                & CE-OCL                       & \textbf{1.00} (0.00)                        & --                                             & \textbf{-12.21} (2.67)                                 & \textbf{0.94} (0.01)                       \\
\rowcolor{gray!10} & CE-OCL\_tr & \textbf{1.00} (0.00) & -- & -92.71 (13.75) & 0.75 (0.02) \\
\rowcolor{gray!10}                & DiCE                         & 0.97 (0.03)                        & --                                             & -203.83 (13.70)                               & 0.22 (0.02)                       \\
\rowcolor{gray!10} HELOC           & Actionable Recourse$^*$ & --                                 & --                                             & --                                            & --                                \\
\rowcolor{gray!10}                & Growing spheres              & 0.77 (0.08)                        & --                                             & -87.36 (13.16)                                & 0.00 (0.00)                       \\
\rowcolor{gray!10}                & FACE                         & 0.77 (0.08)                        & --                                             & -361.80 (25.20)                               & 0.17 (0.02)                       \\ 
                & CE-OCL                       & \textbf{1.00} (0.00)                        & --                                             & \textbf{-1.18} (0.66)                                  & \textbf{0.90} (0.01)                       \\
                & CE-OCL\_tr & \textbf{1.00} (0.00) & -- & -120.61 (118.54) & 0.87 (0.01) \\
                & DiCE                         & \textbf{1.00} (0.00)                        & --                                             & -1618.33 (305.53)                             & 0.25 (0.02)                       \\
 CREDIT          & Actionable Recourse          & 0.83 (0.17)                        & --                                             & -8.47 (7.96)                                  & 0.88 (0.02)                       \\
                & Growing spheres              & 0.63 (0.09)                        & --                                             & -47.73 (27.24)                                & 0.10 (0.00)                       \\
                & FACE                         & 0.63 (0.09)                        & --                                             & -3001.73 (430.61)                             & 0.11 (0.02)                       \\
\bottomrule
\multicolumn{6}{l}{\footnotesize For the comparison, one counterfactual was generated for each of 30 factual instances.} \\ [-2mm]
\multicolumn{6}{l}{\footnotesize The scores were averaged over all instances, and the standard error was derived.} \\  [-2mm] 
\multicolumn{6}{l}{\footnotesize * For the Heloc dataset, Actionable Recourse did not yield any counterfactuals for any of the thirty factual instances.} \\
\end{tabular}%
}
\end{table}

We performed a more thorough comparison between \texttt{CE-OCL} and \texttt{DiCE} on the same four datasets but this time generating three CEs for each instance and using all the predictive models supported by both \texttt{OptiCL} and \texttt{DiCE}. In Table~\ref{tab:ce-ocl_vs_dice}, we report the results obtained with \texttt{CE-OCL} including validity, proximity, coherence, sparsity, diversity, and actionability together with the results obtained considering also the data manifold closeness, (\texttt{CE-OCL\_tr}). The results clearly show how \texttt{CE-OCL} outperforms \texttt{DiCE} in terms of validity, categorical proximity, continuous proximity, and sparsity. While both methods have a categorical diversity score very close to zero in every scenario, \texttt{DiCE} has a generally better performance in terms of continuous diversity. Similarly, \texttt{DiCE} has a better sparsity-based diversity score with the exception of the COMPAS dataset. The addition of closeness constraints (\texttt{CE-OCL}\_tr) has a negative effect on the sparsity and proximity scores but it positively affects the diversity scores when compared to \texttt{CE-OCL}. This was to be expected, as the data manifold region forces solutions to be located in a high-density region, which might lead to optimal solutions with more feature changes. While the sparsity decreases, this loss comes at a high potential of more valuable counterfactuals.

\renewcommand{\baselinestretch}{0.95} 

\begin{table}[ht]
    \caption{Comparison of CE-OCL, CE-OCL with trust region, and DiCE (genetic) with a range of preditive models.}
    

\resizebox{\textwidth}{!}{%
\begin{tabular}{p{0.5cm}p{2cm}p{1.8cm}p{2cm}p{3.7cm}p{2cm}p{2cm}p{3.5cm}p{2.5cm}}
\toprule
        & & \textbf{validity}($\uparrow_0^1$)  &  \textbf{cat. pro-} &  \textbf{cont.} &  \textbf{sparsity}($\uparrow_0^1$) &  \textbf{cat. di-} &  \textbf{cont.} &  \textbf{sparsity-based} \\ [-1mm]
        & &   & \textbf{ximity} ($\uparrow_0^1$)  &  \textbf{proximity}($\uparrow_-^0$)&  & \textbf{versity} ($\uparrow_0^1$) & \textbf{diversity} ($\uparrow_0^+$) &  \textbf{diversity}($\uparrow_0^1$) \\ 
        & & mean (s.e.) & mean (s.e.) & mean (s.e.) & mean (s.e.) & mean (s.e.)  & mean (s.e.) & mean (s.e.) \\ \midrule
\multicolumn{9}{c}{\textbf{Adult dataset}}\\ \midrule
\rowcolor{gray!10}                                      & CE-OCL        & \textbf{1.00} (0.00) & \textbf{0.99} (0.00) & \textbf{-6775.61} (953.24)     & \textbf{0.92} (0.00) & 0.01 (0.01) & 5796.61 (1056.70)    & 0.11 (0.01)        \\
\rowcolor{gray!10} rf                                          & CE-OCL\_tr    & \textbf{1.00} (0.00) & 0.97 (0.02) & -25044.04 (7645.82)   & 0.86 (0.01) & 0.00 (0.00) & 13288.83 (3236.28)   & 0.13 (0.01)        \\
\rowcolor{gray!10}                                          & DiCE & 1.00 (0.00) & 0.74 (0.02) & -81581.97 (8003.12)   & 0.51 (0.01) & \textbf{0.19} (0.02) & \textbf{79595.54} (7496.70)   & \textbf{0.26} (0.02)        \\ 


                                        & CE-OCL        & \textbf{1.00} (0.00) & \textbf{0.99} (0.01) & \textbf{-4226.05} (794.33)     & \textbf{0.89} (0.01) & 0.01 (0.01) & 8042.59 (1418.66)    & 0.19 (0.01)        \\
lr                                          & CE-OCL\_tr    & \textbf{1.00} (0.00) & 0.96 (0.02) & -23288.53 (6837.94)   & 0.84 (0.01) & 0.05 (0.02) & 23421.06 (7376.79)   & 0.22 (0.01)        \\
                                          & DiCE & 0.71 (0.05) & 0.68 (0.03) & -111661.60 (10261.28) & 0.47 (0.02) & \textbf{0.30} (0.03) & \textbf{110668.36} (11019.56) & \textbf{0.35} (0.02)        \\ 
                                          
                                          
\rowcolor{gray!10}                                        & CE-OCL        & \textbf{1.00} (0.00) & 0.88 (0.02) & \textbf{-13994.62} (4198.07)   & 0.83 (0.02) & 0.21 (0.04) & 26889.72 (8343.13)   & \textbf{0.27} (0.03)        \\
\rowcolor{gray!10} cart                                      & CE-OCL\_tr    & \textbf{1.00} (0.00) & \textbf{0.94} (0.01) & -19490.98 (5725.22)   & \textbf{0.84} (0.01) & 0.08 (0.02) & 20313.32 (5702.44)   & 0.19 (0.01)        \\
\rowcolor{gray!10}                                           & DiCE  & 0.65 (0.06) & 0.77 (0.02) & -91507.16 (10271.17)  & 0.55 (0.02) & \textbf{0.23} (0.02) & \textbf{84111.77} (11802.96)  & 0.25 (0.01)        \\ 


                                          & CE-OCL        & \textbf{1.00} (0.00) & \textbf{1.00} (0.00) & \textbf{-10553.10} (1842.00)   & \textbf{0.88} (0.01) & 0.00 (0.00) & 2538.19 (669.37)     & 0.15 (0.02)        \\
mlp                                          & CE-OCL\_tr    & \textbf{1.00} (0.00) & 0.97 (0.02) & -21467.01 (7465.77)   & 0.86 (0.01) & 0.00 (0.00) & 6229.99 (2385.40)    & 0.14 (0.01)        \\
                                          & DiCE & 0.62 (0.06) & 0.67 (0.03) & -89314.87 (10646.04)  & 0.47 (0.02) & \textbf{0.26} (0.03) & \textbf{83848.59} (11476.56)  & \textbf{0.31} (0.02)        
                                          \\ 
                                          
\rowcolor{gray!10}                                          & CE-OCL        & \textbf{1.00} (0.00) & \textbf{1.00} (0.00) & \textbf{-2488.82} (865.72)     & \textbf{0.91} (0.00) & 0.00 (0.00) & 4475.59 (1737.16)    & 0.13 (0.01)        \\
\rowcolor{gray!10} gbm                                          & CE-OCL\_tr    & \textbf{1.00} (0.00) & 0.97 (0.01) & -22732.34 (7545.41)   & 0.88 (0.01) & 0.02 (0.01) & 10609.72 (3583.71)   & 0.15 (0.01)        \\
\rowcolor{gray!10}                                          & DiCE & 0.91 (0.04) & 0.70 (0.02) & -113297.47 (11606.39) & 0.49 (0.01) & \textbf{0.25} (0.02) & \textbf{75728.74} (10042.23)  & \textbf{0.27} (0.02)        \\ \bottomrule
\end{tabular}%
}

\resizebox{\textwidth}{!}{%
\begin{tabular}{p{0.5cm}p{2cm}p{1.8cm}p{2cm}p{3.7cm}p{2cm}p{2cm}p{3.5cm}p{2.5cm}}
\multicolumn{9}{c}{\textbf{COMPAS dataset}}\\ \midrule

\rowcolor{gray!10}                     & CE-OCL     & \textbf{1.00} (0.00) & \textbf{1.00} (0.00) & \textbf{-18.79} (6.15)   & \textbf{0.85} (0.00) & 0.00 (0.00) & 8.87 (4.30)    & 0.17 (0.01) \\
\rowcolor{gray!10} rf                   & CE-OCL\_tr & \textbf{1.00} (0.00) & \textbf{1.00} (0.00) & -38.02 (16.46)  & \textbf{0.85} (0.01) & 0.00 (0.00) & 9.41 (4.49)    & 0.16 (0.01) \\
\rowcolor{gray!10}                     & DiCE       & 0.81 (0.04) & 0.96 (0.01) & -40.94 (7.06)   & 0.60 (0.01) & \textbf{0.06} (0.02) & \textbf{22.30} (5.78)   & \textbf{0.20} (0.01) \\ 


                     & CE-OCL     & \textbf{1.00} (0.00) & \textbf{1.00} (0.00) & -121.74 (13.72) & \textbf{0.80} (0.01) & 0.00 (0.00) & \textbf{229.89} (28.27) & 0.34 (0.01) \\
lr               & CE-OCL\_tr & \textbf{1.00} (0.00) & 0.98 (0.01) & \textbf{-35.84} (11.03)  & 0.75 (0.01) & 0.01 (0.01) & 34.68 (7.79)   & \textbf{0.35} (0.01) \\
                     & DiCE       & 0.85 (0.05) & 0.94 (0.02) & -56.05 (12.83)  & 0.59 (0.01) & \textbf{0.08} (0.02) & 30.61 (9.48)   & 0.21 (0.02) \\ 
                     
                     
\rowcolor{gray!10}                     & CE-OCL     & \textbf{1.00} (0.00) & \textbf{1.00} (0.00) & \textbf{-23.14} (6.03)   & \textbf{0.84} (0.01) & 0.00 (0.00) & \textbf{33.38} (11.83)  & \textbf{0.19} (0.01) \\
\rowcolor{gray!10} cart                 & CE-OCL\_tr & \textbf{1.00} (0.00) & \textbf{1.00} (0.00) & -28.43 (9.16)   & 0.83 (0.01) & 0.00 (0.00) & 31.16 (10.10)  & \textbf{0.19} (0.01) \\
\rowcolor{gray!10}                     & DiCE       & 0.77 (0.08) & 0.96 (0.01) & -32.99 (6.04)   & 0.60 (0.01) & \textbf{0.07} (0.02) & 24.31 (6.63)   & 0.18 (0.01) \\ 


                     & CE-OCL     & \textbf{1.00} (0.00) & \textbf{1.00} (0.00) & \textbf{-16.55} (2.11)   & 0.81 (0.01) & 0.00 (0.00) & 16.20 (4.62)   & \textbf{0.22} (0.01) \\
mlp                  & CE-OCL\_tr & \textbf{1.00} (0.00) & \textbf{1.00} (0.00) & -27.75 (10.25)  & \textbf{0.82} (0.01) & 0.00 (0.00) & 7.29 (4.09)    & 0.18 (0.01) \\
                     & DiCE       & 0.82 (0.06) & 0.96 (0.01) & -59.11 (13.01)  & 0.58 (0.01) & \textbf{0.06} (0.02) & \textbf{24.95} (5.78)   & \textbf{0.22} (0.02) \\ 
                     
                     
\rowcolor{gray!10}                     & CE-OCL     & \textbf{1.00} (0.00) & \textbf{1.00} (0.00) & \textbf{-10.10} (2.60)   & \textbf{0.86} (0.00) & 0.00 (0.00) & 13.64 (3.09)   & \textbf{0.21} (0.01) \\
\rowcolor{gray!10} gbm                  & CE-OCL\_tr & \textbf{1.00} (0.00) & \textbf{1.00} (0.00) & -25.70 (10.66)  & 0.85 (0.01) & 0.00 (0.00) & 13.49 (5.08)   & 0.20 (0.01) \\
\rowcolor{gray!10}                     & DiCE       & 0.59 (0.07) & 0.96 (0.01) & -42.64 (6.32)   & 0.60 (0.01) & \textbf{0.08} (0.02) & \textbf{24.51} (5.81)   & 0.20 (0.01) \\ 
\bottomrule
\end{tabular}%
}


\resizebox{\textwidth}{!}{%
\begin{tabular}{p{0.5cm}p{2cm}p{1.8cm}p{2cm}p{3.7cm}p{2cm}p{2cm}p{3.5cm}p{2.5cm}}
\multicolumn{9}{c}{\textbf{Heloc dataset}}\\ \midrule

\rowcolor{gray!10}                     & CE-OCL     & \textbf{1.00} (0.00) & -- & \textbf{-13.53} (2.35)   & \textbf{0.93} (0.00) & -- & 9.94 (2.74)    & 0.09 (0.01) \\
\rowcolor{gray!10} rf                   & CE-OCL\_tr & \textbf{1.00} (0.00) & -- & -94.24 (13.68)  & 0.75 (0.02) & -- & 18.93 (4.62)   & 0.24 (0.02) \\
\rowcolor{gray!10}                     & DiCE       & 0.90 (0.03) & -- & -231.05 (11.17) & 0.21 (0.02) & -- & \textbf{223.91} (14.16) & \textbf{0.61} (0.02) \\ 
                     

                     & CE-OCL     & \textbf{1.00} (0.00) & -- & \textbf{-99.09} (14.22)  & \textbf{0.88} (0.01) & -- & 188.16 (28.38) & 0.21 (0.01) \\
lr               & CE-OCL\_tr & \textbf{1.00} (0.00) & -- & -138.29 (16.52) & 0.72 (0.02) & -- & 72.51 (8.39)   & 0.34 (0.02) \\
                     & DiCE       & 0.70 (0.06) & -- & -232.39 (12.87) & 0.21 (0.02) & -- & \textbf{207.02} (11.34) & \textbf{0.61} (0.02) \\ 
                     
                     
\rowcolor{gray!10}                     & CE-OCL     & \textbf{1.00} (0.00) & -- & \textbf{-13.12} (1.40)   & \textbf{0.95} (0.00) & -- & 19.72 (2.41)   & 0.08 (0.00) \\
\rowcolor{gray!10} cart                 & CE-OCL\_tr & \textbf{1.00} (0.00) & -- & -99.05 (13.45)  & 0.73 (0.02) & -- & 41.03 (6.53)   & 0.31 (0.02) \\
\rowcolor{gray!10}                     & DiCE       & 0.80 (0.07) & -- & -216.70 (13.47) & 0.22 (0.02) & -- & \textbf{234.89} (16.10) & \textbf{0.61} (0.02) \\ 
                     
                     
                     & CE-OCL     & \textbf{1.00} (0.00) & -- & \textbf{-25.09} (7.57)   & \textbf{0.92} (0.01) & -- & 21.30 (4.18)   & 0.12 (0.01) \\
mlp                  & CE-OCL\_tr & \textbf{1.00} (0.00) & -- & -98.94 (15.87)  & 0.75 (0.02) & -- & 15.41 (5.52)   & 0.26 (0.02) \\
                     & DiCE       & 0.67 (0.07) & -- & -252.56 (14.17) & 0.20 (0.02) & -- & \textbf{246.96} (16.31) & \textbf{0.61} (0.02) \\ 
                     
                     
\rowcolor{gray!10}                     & CE-OCL     & \textbf{1.00} (0.00) & -- & \textbf{-8.41} (2.45)    & \textbf{0.94} (0.00) & -- & 16.31 (4.92)   & 0.10 (0.00) \\
\rowcolor{gray!10} gbm                  & CE-OCL\_tr & \textbf{1.00} (0.00) & -- & -89.91 (14.70)  & 0.76 (0.02) & -- & 18.70 (6.87)   & 0.25 (0.02) \\
\rowcolor{gray!10}                     & DiCE       & 0.73 (0.08) & -- & -234.96 (11.60) & 0.22 (0.02) & -- & \textbf{248.95} (17.34) & \textbf{0.59} (0.02) \\

\bottomrule
\end{tabular}%
}


\resizebox{\textwidth}{!}{%
\begin{tabular}{p{0.5cm}p{2cm}p{1.8cm}p{2cm}p{3.7cm}p{2cm}p{2cm}p{3.5cm}p{2.5cm}}
\multicolumn{9}{c}{\textbf{Give me some credit dataset}}\\ \midrule

\rowcolor{gray!10}                    & CE-OCL     & \textbf{1.00} (0.00) & -- & \textbf{-6.77} (4.43)      & \textbf{0.90} (0.00) & -- & 9.14 (5.53)      & 0.15 (0.01) \\
\rowcolor{gray!10} rf                   & CE-OCL\_tr & \textbf{1.00} (0.00) & -- & -97.01 (95.21)    & 0.89 (0.01) & -- & 115.65 (113.90)  & 0.16 (0.01) \\
\rowcolor{gray!10}                     & DiCE       & \textbf{1.00} (0.00) & -- & -2166.72 (318.36) & 0.23 (0.02) & -- & \textbf{2446.71} (455.17) & \textbf{0.32} (0.01) \\
                     
                     
                     & CE-OCL     & \textbf{1.00} (0.00) & -- & \textbf{-3.79} (1.24)      & \textbf{0.88} (0.01) & -- & 7.50 (2.49)      & 0.24 (0.01) \\
lr               & CE-OCL\_tr & \textbf{1.00} (0.00) & -- &   -614.00 (202.97)           &     0.83 (0.01)        & -- & 1107.84 (381.92)                  &  0.25 (0.01)           \\
                     & DiCE       & 0.92 (0.05) & -- & -1946.86 (256.37) & 0.21 (0.02) & -- & \textbf{1909.26} (187.85) & \textbf{0.29} (0.01) \\ 
                     
                     
\rowcolor{gray!10} & CE-OCL     & \textbf{1.00} (0.00) & -- & -1.85 (0.23)      & \textbf{0.87} (0.00) & -- & 1.91 (0.23)      & 0.17 (0.00) \\
\rowcolor{gray!10} cart                 & CE-OCL\_tr & \textbf{1.00} (0.00) & -- &  -212.82 (100.91)                 &  0.85 (0.01)           & -- &     285.60 (121.56)             &       0.22 (0.01)      \\
\rowcolor{gray!10}                     & DiCE       & 0.00 (0.00) & -- & -1895.95 (230.21) & 0.25 (0.02) & -- & \textbf{2214.51} (319.98) & \textbf{0.32} (0.01) \\
                     
                     
                     & CE-OCL     & \textbf{1.00} (0.00) & -- & \textbf{-24.21} (8.71)     & \textbf{0.89} (0.00) & -- & 38.15 (13.75)    & 0.15 (0.01) \\
mlp                  & CE-OCL\_tr & \textbf{1.00} (0.00) & -- & -996.30 (370.04) & 0.85 (0.01)  & -- &  971.37 (447.54)                &  0.17 (0.01)           \\
                     & DiCE       & 0.97 (0.03) & -- & -2526.22 (265.46) & 0.20 (0.02) & -- & \textbf{3205.58} (427.42) & \textbf{0.32} (0.01) \\
                     
                     
\rowcolor{gray!10}                     & CE-OCL     & \textbf{1.00} (0.00) & -- &    \textbf{-175.98} (74.32)               &    \textbf{0.89} (0.01)         & -- &     296.26 (134.93)             &  0.17 (0.01)           \\
\rowcolor{gray!10} gbm                  & CE-OCL\_tr & \textbf{1.00} (0.00) & -- & -219.19 (131.96)           &     0.87 (0.01)        & -- &    123.61 (82.17)              &    0.16 (0.01)          \\
\rowcolor{gray!10}                     & DiCE       &  0.93 (0.04)           & -- &   -2222.50 (277.89)                &   0.22 (0.02)          & -- &    \textbf{2749.12} (409.75)              &     \textbf{0.31} (0.02)        \\

\bottomrule
\end{tabular}%
}

    \label{tab:ce-ocl_vs_dice}
\end{table}

\clearpage
\section{Case studies}\label{app:case study}

We reserve this appendix for the details of our case study, Statlog (German Credit Data) dataset, and for the additional demonstration on the Statlog (Heart) dataset.

\subsection{German Credit Data tables}\label{app:german_tables}
We report an overview of the Statlog (German Credit Data) dataset features and the CEs generated at each step detailed in Section~\ref{sec:experiments} in Table~\ref{tab:description} and Table~\ref{tab:credit_demo}, respectively. 
\renewcommand{\baselinestretch}{1.5} 

\begin{table}[h]
\caption{Information on Statlog (German Credit Data) Data Set \protect \citep[][]{Dua:2019}}
\label{tab:description}
\resizebox{\textwidth}{!}{%
\begin{tabular}{@{}lllll@{}}
\toprule
\textbf{Label} & \textbf{Variable name} & \textbf{Description} & \textbf{Domain$*$} & \textbf{Constraint} \\ \midrule
F1 & duration & Duration in months & real & $\geq 0$ \\
\rowcolor{gray!10}F2 & credit\_amount & Credit amount & real & $\geq 0$ \\
F3 & instalment\_commitment & Installment rate in percentage of disposable income & real & $\geq 0$ \\
\rowcolor{gray!10}F4 & age & Age in years & real & $x_{age} \geq \hat{\vx}_{age}$ \\
F5 & residence\_since & Present residence since X years & integer  & $x_{residence\_since} \geq \hat{\vx}_{residence\_since}$ \\
\rowcolor{gray!10}F6 & existing\_credits & Number of existing credits at this bank & integer & $\geq 0$ \\
F7 & num\_dependents & Number of people being liable to provide maintenance for & integer & $\geq 0$ \\
\rowcolor{gray!10}F8 & checking\_status & Status of existing checking account, in Deutsche Mark & binary & -- \\
F9 & credit\_history & Credit history (credits taken, paid back duly, delays, critical accounts) & binary & -- \\
\rowcolor{gray!10}F10 & employment & Present employment, in number of years. & binary & conditionally immutable \\
F11 & foreign\_worker & Foreign worker (yes,no) & binary & immutable \\
\rowcolor{gray!10}F12 & housing & Housing (rent, own,...) & binary & -- \\
F13 & job & Job & binary & -- \\
\rowcolor{gray!10}F14 & other\_parties & Other debtors / guarantors & binary & -- \\
F15 & other\_payment\_plans & Other installment plans (banks, stores) & binary & -- \\
\rowcolor{gray!10}F16 & own\_telephone & Telephone (yes,no) & binary & -- \\
F17 & personal\_status & Personal status (married, single,...) and sex & binary & \begin{tabular}[c]{@{}l@{}}immutable\end{tabular} \\
\rowcolor{gray!10}F18 & property\_magnitude & Property (e.g. real estate) & binary & -- \\
F19 & purpose & Purpose of the credit (car, television,...) & binary & \begin{tabular}[c]{@{}l@{}}immutable \end{tabular} \\
\rowcolor{gray!10}F20 & saving\_status & Status of savings account/bonds, in Deutsche Mark. & binary & -- \\ \bottomrule
\multicolumn{5}{l}{\footnotesize $^*$ All categorical are one-hot encoded and therefore considered binary.}
\end{tabular}}
\end{table}
\renewcommand{\baselinestretch}{1.2} 

\begin{table}[ht]
\caption{\protect \texttt{CE-OCL} demo on the Statlog (German Credit Data) Data Set \citep[][]{Dua:2019}.}
\centering
\resizebox{\textwidth}{!}{%
\begin{tabular}{p{1cm}p{1cm}p{1.4cm}p{1cm}p{1cm}p{1cm}p{1cm}p{1cm}p{1.4cm}p{1cm}p{1cm}p{1cm}}

\multicolumn{12}{l}{(a) Counterfactual explanations generated for enriching the optimization model step by step with the} \\
\multicolumn{12}{l}{constraint presented in Section~\ref{sec:generate}} \\

\toprule

{} &     F1 &       F2 &    F3 &     F4 &   F8$^*$ &     F10 &   F12 &        F14 &   F16 & F18$^*$ & F20$^*$ \\
\midrule
\rowcolor{gray!25} $\hat{\bm{x}}$ &   24.0 &  1371.26 &   4.0 &   25.0 &  A &   1$\leq$X$<$4 &  rent &       none &  none &   A &   A \\

\rowcolor{gray!10}\multicolumn{12}{l}{\textbf{Part A}: validity, proximity, coherence} \\

(a)     &  15.02 &	-333.52	& 3.86 &	27.04 &  -- &       -- &     -- &          -- &     -- &   -- &   -- \\

\rowcolor{gray!10}\multicolumn{12}{l}{\textbf{Part B}: validity, proximity, coherence, sparsity} \\

(a)     &   7.12 & -- &     -- &  -- &  -- &       -- &     -- &          -- &     -- &   -- &   -- \\

\rowcolor{gray!10}\multicolumn{12}{l}{\textbf{Part C}: validity, proximity, coherence, sparsity, diversity} \\

(a)     &   7.12 &     -- &     -- &  -- &  -- &       -- &     -- &          -- &     -- &   -- &   -- \\
(b)     &      -- & -2873.47 &     -- &      30.06 &  -- &       -- &     -- &          -- &     -- &   -- &   -- \\
(c)     &      -- &        -- &     1.96 &      26.63 & -- &       -- &     -- &          -- &     -- &   -- &   -- \\

\rowcolor{gray!10}\multicolumn{12}{l}{\textbf{Part D}: validity, proximity, coherence, sparsity, diversity, actionability} \\

(a)     &   7.12 &  -- &     -- &  -- &  -- &       -- &     -- &  -- &     -- &   -- &   -- \\
(b) & -- &  -- &  1.96 &  26.63 &  -- &  -- &     -- &  -- &     -- &   -- &   -- \\
(c) &  -- &  -- &  -- &  75.52 &   -- &  -- &     -- &  -- &  -- &   -- &   -- \\

\rowcolor{gray!10}\multicolumn{12}{l}{\textbf{Part E}: validity, proximity, coherence, sparsity, diversity, actionability, data manifold closeness} \\

(a)     &   22.0 &  1283.52 &     -- &      -- &  B &   4$\leq$X$<$7 &     -- &          -- &     -- &   B &   -- \\
(b)     &      10 &  1363.43 &     2.0 &   64.0 &  B &       -- &     own &          -- &   yes &   C &   B \\
(c)     &   12.0 &  1893.04 &     -- &   29.0 &  -- &       -- &   own &  guarantor &   yes &   B &   B \\

\rowcolor{gray!10}
\multicolumn{12}{l}{\textbf{Part F}: validity, proximity, coherence, sparsity, diversity, actionability, data manifold closeness, causality} \\
(a)     &      -- &        -- &     -- &      --  &  B &  4<=X<7 &     -- &          -- &     -- &   B &   -- \\
(b)     &   22.0 &   990.51 &     -- &      -- &  B &   4<=X<7 &     -- &          -- &     -- &   B &   -- \\
(c)     &  26.83 &  1910.28 &     -- &      -- &  B &  4<=X<7 &     -- &          -- &     -- &   B &   -- \\

\bottomrule

\multicolumn{12}{l}{\footnotesize F1--F20 represent the 20 features of the dataset. See Table \ref{tab:description} in Appendix \ref{app:case study} for a description.} \\[-2mm]
\multicolumn{12}{l}{\footnotesize The dash (--) represents no change in a feature with respect to the factual instance.} \\[-2mm]
\multicolumn{12}{l}{\footnotesize\textbf{F5, F6, F7, F9, F11, F13, F15, F17, F19:} None of the counterfactual explanations proposed a change in these variables. For space }\\[-2mm]
\multicolumn{12}{l}{\footnotesize reasons they are not displayed here.} \\[-2mm]
\multicolumn{12}{l}{\footnotesize $^*$ \textbf{F8:} A: $$<$$0, B: no checking; \textbf{F18:} A: real estate, B: life insurance, C: car ; \textbf{F20:} A: no known savings, B: $$<$$100} \\
\end{tabular}}

\vspace{0.4cm}

        \centering
        \resizebox{\textwidth}{!}{%
        \begin{tabular}{lcccccc}
        
        \multicolumn{7}{l}{\normalsize{(b) Evaluation$^*$ of counterfactuals generated for a single factual instance, with constraints added gradually.}} \\

        \toprule
        
        {} &  \small \textbf{categorical} &  \small\textbf{continuous} &  \small\textbf{sparsity}($\uparrow_0^1$) &  \small\textbf{categorical} &  \small\textbf{continuous} &  \small\textbf{sparsity-based} \\ [-2mm]
        
        {} &  \small\textbf{proximity}($\uparrow_0^1$) &  \small\textbf{proximity}($\uparrow_-^0$) &   & \small \textbf{diversity}($\uparrow_0^1$) &  \small\textbf{diversity}($\uparrow_0^+$) &  \small\textbf{diversity}($\uparrow_0^1$) \\
        
        \midrule
\small Part A & \small               1.00 & \small  -1715.94 &   \small   0.8 &        -- &         -- &                -- \\
\rowcolor{gray!10} \small Part B &              \small  1.00 &    \small  -16.88 &    \small   0.95 &        -- &         -- &                -- \\
\small Part C &             \small   1.00 &  \small  -1423.45 &    \small   0.92 &    \small    0.00 &    \small  2845.81 &         \small       0.15 \\
\rowcolor{gray!10}\small Part D &            \small    1.00 & \small     -23.69 &  \small     0.93 &     \small   0.00 &   \small     46.29 &   \small             0.12 \\
\small Part E &        \small        0.67 &   \small  -230.12 &    \small   0.63 &     \small   0.36 &     \small  441.68 & 0.42 \\
\rowcolor{gray!10}\small Part F &       \small         0.77 & \small    -308.20 &   \small    0.78 &      \small  0.00 &   \small    616.40 &        \small        0.10 \\

        \bottomrule
\multicolumn{7}{l}{\footnotesize \textbf{Part A:} validity, proximity, coherence; \textbf{Part B:} validity, proximity, coherence, sparsity; \textbf{Part C:} validity, proximity, coherence, } \\[-2mm]
\multicolumn{7}{l}{\footnotesize sparsity, diversity; \textbf{Part D:} validity, proximity, coherence, sparsity, diversity, actionability; \textbf{Part E:} validity, proximity, coherence, } \\[-2mm]
\multicolumn{7}{l}{\footnotesize sparsity, diversity, actionability, data manifold closeness; \textbf{Part F:} validity, proximity, coherence, sparsity, diversity, actionability.} \\ [-2mm]
\multicolumn{7}{l}{\footnotesize data manifold closeness, causality} \\ [-2mm]
\multicolumn{7}{l}{\footnotesize \textbf{$^*$validity ($\uparrow_0^1$):} 1.00 in all cases } \\ 
      \end{tabular}}
       \label{tab:credit_demo}

\end{table}

\clearpage
\subsection{German Credit Data CE generation model}\label{app:german_model}
For the case study in Section~\ref{sec:experiments}, we made use of the Statlog (German Credit Data) dataset \citep[][]{Dua:2019} \footnote{Preprocessed from \protect  \hyperlink{https://datahub.io/machine-learning/credit-g}{https://datahub.io/machine-learning/credit-g}.}. Table~\ref{tab:description} provides an overview of features in this dataset, alongside a short description and the measurement level. For conciseness, we labelled the features F1-F20, and use those labels throughout the manuscript. Table~\ref{tab:description} also displays the actionability constraints we imposed on the features.
The following  mathematical model is used to generate CEs and contains all the constraints -- criteria -- presented in Section~\ref{sec:generate}:

\allowdisplaybreaks
\begin{subequations}
\begin{align}
\underset{{\bm{x}, \bm{z}, \bm{s}\in \mathbb{R}^n, \bm{\lambda} \in \mathbb{R}_{\geq0}^{|\mathcal{I}|}}}{\minimize} \ & \ell_2(\bm{x}, \bm{\hat{x}}) + \alpha \sum_i z_i + \beta \ell_1(\bm{s}, \bm{\tilde{s}})  & \owntag[eq:final0]{proximity, sparsity, and closeness}\\
\subto \ & h(\bm{x}) = 1 \owntag[eq:final1]{validity}\\
& |\bm{x} - \bm{\hat{x}}| \leq M \bm{z},\owntag[eq:final2]{sparsity}\\
& \sum_{i \in \mathcal{I}} \lambda_i \bm{\bar{x}}_i = \bm{x} + \bm{s}, & \owntag[eq:final3]{data manifold closeness}\\
& \sum_{i \in \mathcal{I}} \lambda_i = 1, &\owntag[eq:final4]{data manifold closeness}\\
& x_i \geq 0, ~~ i \in \{F1, F2, F3, F6 ,F7\} \owntag[eq:final5]{actionability}\\
& x_i \geq \hat{x}_i, ~~ i \in \{F4, F5\}\owntag[eq:final6]{actionability}\\
& x_i = \hat{x}_i, ~~ i \in \{F11, F17, F19\}\owntag[eq:final7]{immutability}\\
& x_{F10} \in \CC_{F10},\owntag[eq:final8]{conditional immutability}\\
& x_{F1} = \hat{x}_{F1} + h_{causality}(x_{F2}) - h_{causality}(\hat{x}_{F2}), \owntag[eq:final9]{causality}\\
& \bm{x} \in \mathcal{L}, \owntag[eq:final10]{Domain (real, integer, binary)}
\end{align}
\end{subequations}

\clearpage
\subsection{Heart tables}\label{app:heart_tables}
Similarly to the German Credit Data case study, we report Table~\ref{tab:heart_demo} with the CEs generated at each step and the scores for the evaluation metrics, and Table~\ref{tab:description-heart} with an overview of the features.
\renewcommand{\baselinestretch}{1.5} 

\begin{table}[h]
\caption{CE-OCL demo on the Statlog (Heart) Data Set \citep[][]{Dua:2019}}
\centering
\resizebox{\textwidth}{!}{
\begin{tabular}{lccccccccccccc}
\multicolumn{14}{l}{\LARGE (a) Counterfactual explanations generated for enriching the optimization model step by step with the} \\
\multicolumn{14}{l}{\LARGE constraint presented in Section~\ref{sec:generate}}\\
\toprule
 & \multicolumn{1}{c}{\cellcolor[HTML]{FFFFFF}age} & \multicolumn{1}{c}{\cellcolor[HTML]{FFFFFF}bp} & \multicolumn{1}{c}{\cellcolor[HTML]{FFFFFF}sch} & \multicolumn{1}{c}{\cellcolor[HTML]{FFFFFF}mhrt} & \multicolumn{1}{c}{\cellcolor[HTML]{FFFFFF}opk} & \multicolumn{1}{c}{\cellcolor[HTML]{FFFFFF}chp} & \multicolumn{1}{c}{\cellcolor[HTML]{FFFFFF}ecg} & \multicolumn{1}{c}{\cellcolor[HTML]{FFFFFF}exian} & \multicolumn{1}{c}{\cellcolor[HTML]{FFFFFF}fbs} & \multicolumn{1}{c}{\cellcolor[HTML]{FFFFFF}sex} & \multicolumn{1}{c}{\cellcolor[HTML]{FFFFFF}slope} & \multicolumn{1}{c}{\cellcolor[HTML]{FFFFFF}thal} & \multicolumn{1}{c}{\cellcolor[HTML]{FFFFFF}vessel} \\ \midrule
\rowcolor{gray!25}
$\bm{\hat{x}}$ & 49.0 & 130.0 & 265.98 & 171.01 & 0.6 & atypical angina & normal & no & false & male & upsloping & normal & 0 \\
\rowcolor{gray!10} 
\multicolumn{14}{l}{\cellcolor[HTML]{EFEFEF}\textbf{Part A}: validity, proximity, coherence} \\

(a) & 48.82 & 139.28 & 328.09 & 153.98 & 1.05 & - & - & - & - & - & - & - & - \\
\rowcolor{gray!10} 
\multicolumn{14}{l}{\cellcolor[HTML]{EFEFEF}\textbf{Part B}: validity, proximity, coherence, sparsity} \\

(a) & - & - & 407.24 & - & - & - & - & - & - & - & - & - & - \\
\rowcolor{gray!10} 
\multicolumn{14}{l}{\cellcolor[HTML]{EFEFEF}\textbf{Part C}: validity, proximity, coherence, sparsity, diversity} \\

(a) & - & - & 407.24 & - & - & - & - & - & - & - & - & - & - \\

(b) & - & - & 393.92 & 175.14 & - & - & - & - & - & - & - & - & - \\

(c) & - & - & 404.04 & - & 0.47 & - & - & - & - & - & - & - & - \\
\rowcolor{gray!10} 
\multicolumn{14}{l}{\cellcolor[HTML]{EFEFEF}\textbf{Part D}: validity, proximity, coherence, sparsity, diversity, actionability} \\

(a) & - & - & 407.24 & - & - & - & - & - & - & - & - & - & - \\

(b) & - & - & 393.92 & 175.14 & - & - & - & - & - & - & - & - & - \\

(c) & - & - & - & 124.37 & - & - & - & - & - & - & - & - & - \\
\rowcolor{gray!10} 
\multicolumn{14}{l}{\cellcolor[HTML]{EFEFEF}\textbf{Part E}: validity, proximity, coherence, sparsity, diversity, actionability, data manifold closeness} \\

(a) & - & 111.77 & 253.81 & 152.7 & 0.0 & nonanginal pain & - & - & - & - & - & - & - \\

(b) & - & 137.0 & 258.5 & 147.01 & 1.55 & asymptomatic & left ventricular hypertrophy & - & - & - & flat & reversible defect & - \\

(c) & - & 140.61 & 274.7 & 128.61 & 0.49 & asymptomatic & left ventricular hypertrophy & yes & - & - & - & reversible defect & - \\ \bottomrule

\multicolumn{14}{l}{See Table \ref{tab:description-heart}  for a description of each feature} \\[-2mm]
\multicolumn{14}{l}{The dash (--) represents no change in a feature with respect to the factual instance.} \\[-2mm]

\end{tabular}
}

\vspace{0.4cm}

        \centering
        \resizebox{\textwidth}{!}{%
        \begin{tabular}{lcccccc}
        
        \multicolumn{7}{l}{\normalsize{(b) Evaluation$^*$ of counterfactuals generated for a single factual instance, with constraints added gradually.}} \\

        \toprule
        
        {} &  \small \textbf{categorical} &  \small\textbf{continuous} &  \small\textbf{sparsity}($\uparrow_0^1$) &  \small\textbf{categorical} &  \small\textbf{continuous} &  \small\textbf{sparsity-based} \\ [-2mm]
        
        {} &  \small\textbf{proximity}($\uparrow_0^1$) &  \small\textbf{proximity}($\uparrow_-^0$) &   & \small \textbf{diversity}($\uparrow_0^1$) &  \small\textbf{diversity}($\uparrow_0^+$) &  \small\textbf{diversity}($\uparrow_0^1$) \\

        \midrule
        \rowcolor{gray!10}
        \small Part A  & \small 1.00 & \small -89.05 & \small 0.62 & - & - & - \\
        \small Part B  & \small 1.00 &\small  -141.26 & \small 0.92 & - & - & - \\
        \rowcolor{gray!10}
        \small Part C  & \small 1.00 & \small -137.17 & \small 0.87 & \small 0.00 & \small 11.72 & \small 0.18 \\ 
        \small Part D  & \small  1.00 & \small -106.66 & \small 0.90 & \small 0.00 & \small 128.02 & \small 0.15 \\
        \rowcolor{gray!10} 
        \small Part E  & \small 0.62 & \small -50.19 & \small 0.46 & \small 0.42 & \small 50.25 & \small 0.56\\ \bottomrule
        
\multicolumn{7}{l}{\footnotesize \textbf{Part A:} validity, proximity, coherence; \textbf{Part B:} validity, proximity, coherence, sparsity; \textbf{Part C:} validity, proximity, coherence, } \\[-3mm]

\multicolumn{7}{l}{\footnotesize sparsity, diversity; \textbf{Part D:} validity, proximity, coherence, sparsity, diversity, actionability; \textbf{Part E:} validity, proximity, coherence, } \\[-3mm]

\multicolumn{7}{l}{\footnotesize sparsity, diversity, actionability, data manifold closeness; \textbf{Part F:} validity, proximity, coherence, sparsity, diversity, actionability.} \\ [-3mm]

\multicolumn{7}{l}{\footnotesize data manifold closeness, causality} \\ [-3mm]
        \multicolumn{7}{l}{\footnotesize \textbf{$^*$validity ($\uparrow_0^1$):} 1.00 in all cases}\\
      \end{tabular}}
       \label{tab:heart_demo}

\end{table}
\renewcommand{\baselinestretch}{1.1} 

\begin{table}[ht]
\centering
\caption{Information on Statlog (Heart) Data Set \protect \citep[][]{Dua:2019}}
\label{tab:description-heart}
\resizebox{0.6\textwidth}{!}{%
\begin{tabular}{llll}
\toprule
\textbf{Variable name} & \textbf{Description} & \textbf{Domain$*$} & \textbf{Constraint} \\ \midrule
\rowcolor{gray!10}age & Patient age in years & real & immutable \\
sex & Gender & binary & immutable \\
\rowcolor{gray!10}chp & Chest pain type & binary & - \\
bp & Resting blood pressure & real & $\geq 0$ \\
\rowcolor{gray!10}sch & Serum cholesterol & real & $\geq 0$ \\
fbs & Fasting blood sugar \textgreater 120 mg/dL & binary & - \\
\rowcolor{gray!10}ecg & Resting electrocardiographic result & binary & - \\
mhrt & Maximum heart rate & real & $\geq 0$ \\
\rowcolor{gray!10}exian & Exercise induced angina & binary & - \\
opk & Old peak & real & $\geq 0$ \\
\rowcolor{gray!10}slope & Slope of peak exercise ST segment & binary & - \\
vessel & Number of major  vessels & binary & - \\
\rowcolor{gray!10}thal & Defect type & binary & - \\ \bottomrule
\multicolumn{4}{l}{\footnotesize $^*$ All categorical are one-hot encoded and therefore considered binary.}
\end{tabular}}
\end{table}
\subsection{Heart CE generation model}\label{app:heart_model}
The following  mathematical model is used to generate CEs and contains all the constraints -- criteria -- presented in the Section~\ref{sec:generate}:
\begin{subequations}
\begin{align}
\underset{{\bm{x}, \bm{z}, \bm{s}\in \mathbb{R}^n, \bm{\lambda} \in \mathbb{R}_{\geq0}^{|\mathcal{I}|}}}{\minimize} \ & \ell_2(\bm{x}, \bm{\hat{x}}) + \alpha \sum_i z_i + \beta \ell_1(\bm{s}, \bm{\tilde{s}})  & \owntag[eq:final2_0]{proximity, sparsity, and closeness}\\
\subto \ & h(\bm{x}) = 1 \owntag[eq:final2_1]{validity}\\
& |\bm{x} - \bm{\hat{x}}| \leq M \bm{z},\owntag[eq:final2_2]{sparsity}\\
& \sum_{i \in \mathcal{I}} \lambda_i \bm{\bar{x}}_i = \bm{x} + \bm{s}, & \owntag[eq:final2_3]{data manifold closeness}\\
& \sum_{i \in \mathcal{I}} \lambda_i = 1, &\owntag[eq:final2_4]{data manifold closeness}\\
& x_i \geq 0, ~~ i \in \{bp, sch, mhrt, opk\} \owntag[eq:final2_5]{actionability}\\
& x_i = \hat{x}_i, ~~ i \in \{age, sex\}\owntag[eq:final2_7]{immutability}\\
& \bm{x} \in \mathcal{L}, \owntag[eq:final2_10]{Domain (real, binary)}
\end{align}
\end{subequations}

The predictive model used for this demo is a neural network with one hidden layer of 50 nodes and ReLU activation functions. A description of the Statlog (Heart) dataset used in the experiment is given in Table \ref{tab:description-heart}. The experiments have the same structure described in Section~\ref{sec:experiments}, and the results are reported in Table~\ref{tab:heart_demo}.

\end{document}